\documentclass[conference]{IEEEtran}
\usepackage{authblk}
\usepackage[noadjust]{cite}
\IEEEoverridecommandlockouts
\usepackage{subcaption}
\usepackage{cite}
\usepackage{amsmath,amssymb,amsfonts}
\usepackage{algorithmic}
\usepackage{graphicx}
\usepackage{textcomp}
\usepackage{float}
\usepackage{caption}
\usepackage{multirow}
\usepackage[table,xcdraw]{xcolor}
\usepackage{booktabs}

\def\BibTeX{{\rm B\kern-.05em{\sc i\kern-.025em b}\kern-.08em
 \kern-.1667em\lower.7ex\hbox{E}\kern-.125emX}}
\begin{document}


\title{\LARGE Evaluating the Efficacy of Foundational Models: Advancing Benchmarking Practices to Enhance Fine-Tuning Decision-Making. \\
}


\author{\large Oluyemi Enoch Amujo}
\author{\large Shanchieh Jay Yang}
\affil{\small Rochester Institute of Technology, NY. \\ {oa6121@rit.edu, jay.yang@rit.edu}}

\maketitle

\begin{abstract} 
Recently, large language models (LLMs) have expanded into various domains. However, there remains a need to evaluate how these models perform when prompted with commonplace queries compared to domain-specific queries, which may be useful for benchmarking prior to fine-tuning for domain-specific downstream tasks. This study evaluates LLMs, specifically Gemma-2B and Gemma-7B, across diverse domains, including cybersecurity, medicine, and finance, compared to common knowledge queries. This study utilizes a comprehensive methodology to assess foundational models, which includes problem formulation, data analysis, and the development of ThroughCut, a novel outlier detection technique that automatically identifies response throughput outliers based on their conciseness. This methodological rigor enhances the credibility of the presented evaluation frameworks. This study focused on assessing inference time, response length, throughput, quality, and resource utilization and investigated the correlations between these factors. The results indicate that model size and types of prompts used for inference significantly influenced response length and quality. In addition, common prompts, which include various types of queries, generate diverse and inconsistent responses at irregular intervals. In contrast, domain-specific prompts consistently generate concise responses within a reasonable time. Overall, this study underscores the need for comprehensive evaluation frameworks to enhance the reliability of benchmarking procedures in multidomain AI research.

\end{abstract}


\section{Introduction}
\noindent The increasing efficacy of large language models (LLMs) for dynamic natural language processing (NLP) tasks has attracted significant attention across various disciplines. Notably, LLMs often undergo an intricate development process encompassing training, optimization, fine-tuning, and, in some cases, incorporation of sophisticated techniques like reinforcement learning with human feedback (RLHF), to achieve advanced conversational capabilities. Extensive studies \cite{raiaan2024review}, \cite{fan2023bibliometric} have unveiled the transformative potential of LLMs across various scientific and technological domains. Moreover, their potential to positively impact diverse fields, including software engineering, via process and outcome optimization has been recognized \cite{hou2023large}. Further, their impact on the evolution of artificial intelligence and the burgeoning field of digital learning has been well-documented \cite{DBLP-journals-corr-abs-2308-04889}\cite{doi-10-1080-21532974-2022-2107321}. In addition, LLMs are increasingly being acknowledged for their transformative potential in the cybersecurity field, particularly in the area of threat detection \cite{ferrag2020deep}.

The functions executed by Large Language Models (LLMs) can be fundamentally classified into two categories: input classification and output generation. Tasks falling under the former may encompass Text classification\cite{biswas2023function}, named entity recognition (NER) \cite{dai2019using}, Sentiment Analysis \cite{miah2024multimodal}, question answering \cite{tan2023can} , \cite{arefeen2024leancontext}, and various others. Conversely, output generation tasks may include multimodal output generation  \cite{qu2023layoutllm}, \cite{acharya2023llm}, language understanding and translation \cite{nam2024using},\cite{li2024eliciting}, text summarization \cite{keswani2024abstractive}, dialog systems \cite{goel2023llm}\cite{hu2023unlocking}, among others. Notably, output generation tasks distinguish LLMs from traditional machine learning and deep learning methodologies due to their objective, which is similar to human-like natural language processing tasks.

A significant milestone in artificial intelligence (AI) is the development of large foundational models (LLFMs), which are LLMs pre-trained on diverse datasets across various domains. LLFMs represent substantial progress in AI, offering enhanced performance, broader applicability, and efficiency gains \cite{Jia202110SA}. LLFMs serve as fundamental pillars for numerous natural language processing (NLP) tasks and applications, and they excel at tasks such as text generation, summarization, and translation with exceptional accuracy and fluency. A plethora of LLFMs, varying in size and architecture, have been recently released, including models like Alpaca \cite{alpaca}  \cite{alpaca-2}, BERT, GPT, DALL-E, LLAMA, BLOOM, Gemma, and Alpaca. These models have been developed for diverse applications, spanning multimodal capabilities, vision-and-language tasks, visual-audio processing, as well as code and image generation, among others \cite{singh2022flava}\cite{li2023audio}\cite{li2024multimodal}. By leveraging large datasets for training, these models demonstrate remarkable adaptability, requiring relatively fewer data to fine-tune specific tasks within distinct domains. Thus, they serve as foundational frameworks for many AI applications.

Focusing on the cybersecurity domain, previous studies have explored the potential of fine-tuning LLFMs for various tasks like cyber threat intelligence (CTI) and automation, using natural language text \cite{10.1007/978-3-031-25538-0_3}, identification of intricate patterns for automating software vulnerability detection \cite{Ferrag2023SecureFalconTN}, and many others. However, these studies have revealed a notable inability to evaluate the LFM's baseline cybersecurity knowledge prior to fine-tuning. This gap is exacerbated by the absence of comprehensive pre-fine-tuning assessments, which leads to persistent inaccuracies in benchmarking. In addition, existing evaluation benchmark datasets ROUGE \cite{lin2005recall}, Super-NaturalInstructions \cite{wang2022super}, MMMLU \cite{hendrycks2020measuring} in the realm of LLMs and natural language processing (NLP), inadequately encompass essential cybersecurity materials, which hinders accurate evaluations. Consequently, the misguided adoption of fine-tuning methodologies resulted in flawed benchmarking outcomes (Author, Year). Addressing these issues highlights the urgent need for a more rigorous and comprehensive evaluation framework to rectify these shortcomings and improve the reliability of benchmarking procedures in cybersecurity-AI research.

Therefore, the goal of this study is to evaluate the foundational understanding of special domains such as cybersecurity, finance, and health/medicine, within large foundational models, which facilitates the development of a fine-tuning framework tailored to cybersecurity tasks. Our motivation stems from the intuition that a large language model (LLM) reacts differently to the prompts of various domains. For example, it responds differently to a common query like "What is the meaning of a good life?" compared to a cybersecurity query such as "What is the attacker trying to achieve when running a DLL remotely on the server?". Leveraging this insight, we address inquiries regarding the assessment of foundational model comprehension in various domains such as cybersecurity, finance, and medical. Furthermore, this study is significant because it establishes a framework for developers and researchers to assess the need for fine-tuning. 

\textbf{Our work and findings}. Large language model (LLM) inference requires significant resources, including time, CPU, and memory. Logically, a foundational model trained on a diverse dataset is inclined to retain various forms of knowledge. In our investigation, we observed that their performance in terms of both output and resource usage varies depending on whether they are prompted with a common or domain-specific query.

Summary of our key findings:
\begin{enumerate}
\item 7B models consume more GPU memory than 2B models.
\item Overall, common prompts tend to produce responses with greater diversity in length and longer inference times.
\item Across all categories, the 2B model tends to have higher throughput than its 7B counterpart.
\item There is a strong correlation between inference time and response length compared to the other parameters.
\item When using semantic textual similarity (STS) with ChatGPT responses as a reference, the 7B model exhibits superior performance compared to 2B.
\item 7B model with a response length limit of 50 yields responses with higher ROUGE-L scores in all domains compared to any other parameter.
\end{enumerate}
 
Our contributions are as follows:
\begin{enumerate}
\item We delve into foundation models of diverse sizes, specifically Gemma-2B and Gemma-7B, within both the domain-specific (cybersecurity, health/medical, finance) and common prompt and response generation control settings.
\item Our analysis compares resource utilization and the quality and length of responses generated by the models.
\item We introduce a framework to facilitate informed decision-making when fine-tuning large language models (LLMs) in the cybersecurity, medical, and finance domains.
\item We propose a novel outlier detection technique, termed ThroughCut, which automatically identifies response throughput outliers by assessing their conciseness.

\end{enumerate}

\section{Literature Review}
\noindent In this section, we delve into several concepts crucial to this study, such as the large language foundation model (LLFM), LLM inference, and LLM evaluation metrics.

\subsection{\textbf{Large Language Foundation Model (LLFM)}}
\noindent For the sake of precision and lucidity, the term "foundation models" is employed within the machine learning paradigm antecedent to the emergence of large language models (LLMs), delineating a broader category of AI models that served as a benchmark for user applications \cite{divyanshu-fmos}. 
Moreover, an LLFM, alternatively denoted as a Pre-trained Language Model (PLM) \cite{chiang-etal-2022-recent}\cite{li2024pre}, undergoes training on a comprehensive and diverse dataset to function as a versatile substrate for various applications. After this phase, the model can be fine-tuned on reduced data to perform specific tasks \cite{lu2023framework}. It is important to acknowledge that the capacity and diversity of the foundation model are contingent on the size of the training dataset \cite{10445729}. Therefore, while all LLMs can be categorized as foundation models, not all foundation models attain the scale of largeness.

One common factor among all Large Language and Large Language-Focused Models (LLLFMs) is their development by companies with substantial resources and workforces. These entities include OpenAI, Google Research, MetaAI, and others. For example, GPT-1 was trained using 4.5 GB of text over 30 days on 8 P600 GPUs, equivalent to 1 petaFLOP/s-day, and was publicly released in 2018 \cite{radford2018improving}. In 2023, GPT-4 underwent training involving both text prediction and Reinforcement Learning Hyperparameter Fine-Tuning (RLHF), where the specifics of the data volume and training duration undisclosed, yet estimated to range from 2.1 to 25 FLOP. In addition, more than 50 experts were engaged solely for adversarial testing, in addition to undisclosed others contributing to various facets of the system \cite{aigpt}\cite{KALYAN2024100048}. Llama 3, as described in its model card, was trained using two custom-built 24K GPU clusters, consuming 7.7 million GPU hours and processing over 15 trillion tokens. This dataset is seven times more extensive than the training dataset for Llama 2 \cite{llama3modelcard}\cite{touvron2023llama}.

Fine-tuning is essential when adapting a large language model (LLM) to downstream tasks. There exist various categories of fine-tuning techniques that are worth mentioning. First, fine-tuning the pre-trained parameters can be performed in either a full \cite{lv2023full} or partial \cite{liao-etal-2023-parameter} manner, aiming to update the pre-trained parameters to suit a new task. Although this approach has demonstrated remarkable performance, particularly in domain-specific tasks, it is computationally expensive. Second, parameter-efficient fine-tuning (PEFT) involves adding a small trainable parameter for fine-tuning. PEFT utilizes only a small percentage of existing fine-tuned parameters, referred to as low-rank, to adapt to a downstream task and incorporates them into the pre-trained model \cite{hu2021lora,dettmers2023qlora,chen2024longlora}. While this strategy balances performance and resource efficiency better than full fine-tuning, it increases model size. Finally, prompt-based fine-tuning \cite{gao2023prompt,liu2022p} is a method to construct prompts in a more insightful manner to optimize the model's performance without altering its parameters. In addition, advanced prompt tuning techniques, such as retrieval augmented generation (RAG), have been introduced and demonstrated to effectively mitigate LLM hallucinations \cite{jiawei-rag}. However, a drawback of prompt tuning is that it requires users to have more experience in creating prompts or crafting RAGs that align with their objectives.

In general, the perspective on LLM fine-tuning may vary depending on the researcher’s objectives. A large organization with abundant computing resources may prioritize high-accuracy downstream tasks or specific tasks. Conversely, for a small organization, institution, or individual researcher with limited resources, the objectives may include reducing fine-tuning computational overhead while enhancing overall performance.

\subsection{\textbf{LLM Text Generation and Inference}}
Large language models (LLMs) excel at comprehending human language and extracting insights from corpora of training data. Recent advances in this domain have reached a level of sophistication where distinguishing between machine-generated text and human-authored text has become increasingly challenging, despite numerous investigations \cite{Mo_Qin_Dong_Zhu_Li_2024} \cite{10.1145/3624725}. The text generation task is formally delineated as stated in \cite{10.1145/3649449}.

\begin{equation}
\textit{y}=fM(x,P)
\end{equation}

Here, the text generation model $\text{fM}$ produces the output text $\text{y}$ given the input data $\text{x}$ that satisfies some special set of properties $\text{P}$. The property may be that the input is text, image, tabular data, a knowledge base, etc.

During inference, the text generation model \( M \), typically the decoder, produces output sequences \( y_{i} \) conditionally on some information \( x \), referred to as the prompt, where each \( y_{i} \) represents a token (a word or a subword). Formally, given \( x_{k} \in X \) to a model \( M \) where \( i=1, 2, \ldots, n \), the objective is to predict \( y_{k} \in Y \) where \( j=1, 2, \ldots, k \). The conditional probability denotes this as follows:
\begin{equation}
P(y|x)=P(y_{\text{1}}|x) P(y_{\text{2}}|x, y_{\text{1}} ) \textellipsis P(y_{\text{k}}|x, y_{\text{1}}, \textellipsis,  y_{\text{k-1}}))\label{eq}
\end{equation}
In this context, the model operates in an autoregressive manner \cite{yangxlnet}, generating $y_{\text{i}}$ sequentially and appending it to the target input sequence to predict the subsequent sequence. Consequently, the data structure housing the computational weight mirrors that of a lower triangular matrix \cite{Li2013TriangularMA}\cite{Li2019TriangularizationOM}. Various strategies have been proposed to optimize the value of \( k \), resulting in a lengthy and coherent passage \cite{tan2021progressive}. However, it is essential to maintain a balance between response length and throughput, especially concerning resource-constrained devices.

\subsection{\textbf{Google Gemma Architecture}}
In this subsection, our focus is on Gemma, which serves as a case study for large language foundation models (LLFMs). Google DeepMind released the model in two variations: one with 2 billion parameters and another with 7 billion parameters, as part of the Gemini model series \cite{sundar-demis-gemini}. Although the specific architecture remains undisclosed in the documentation \cite{team2024gemma}, essential components are illustrated in the architecture depicted in Figure 1.

\begin{figure}[ht]
 \centering
 \includegraphics[scale=0.8]{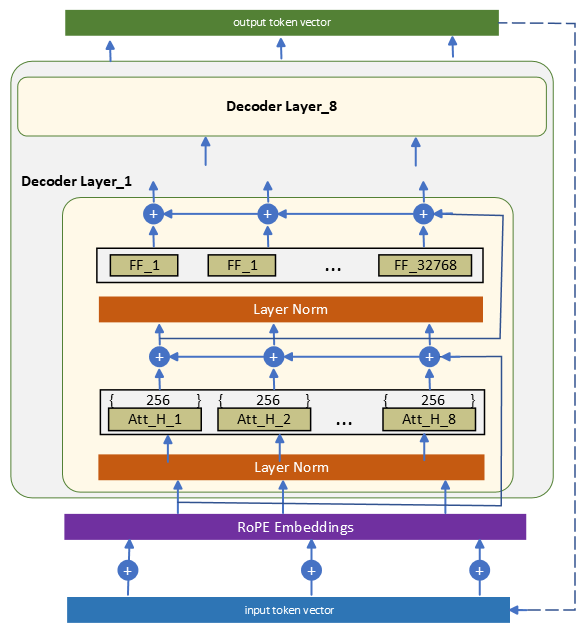}
 \caption{A Gemma-2B architecture showing the salient components}
\end{figure}

The model under consideration constitutes an advanced transformer-decoder \cite{vaswani2017attention} and employs sequence-to-sequence learning techniques  \cite{sutskever2014sequence}. These models have been extensively examined in previous studies, and their autoregressive nature is primarily focused on output generation rather than classification tasks. Essential parameters pertinent to the model are delineated in Figure 2.

\begin{figure}[ht]
 \centering
 \includegraphics[scale=0.5]{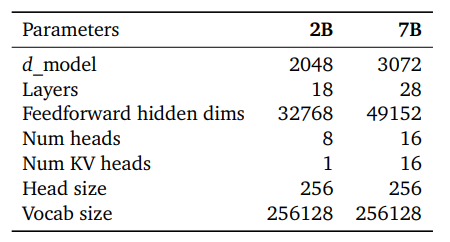}
 \caption{Salient parameters of Gemma model. \textit{Source: \cite{team2024gemma} }}
\end{figure}
Additionally, an essential aspect worth noting regarding Gemma is the Rotary Position Embedding (RoPE) \cite{su2024roformer}. RoPE assigns a position to each token in the input sequence, ensuring accurate positioning in the output sequence. This method considers valuable properties such as sequence length flexibility and decaying inter-token dependency with increasing relative distances. Previous models like BERT \cite{devlin-etal-2019-bert}, GPT \cite{radford2018improving}, and ELECTRA \cite{Clark2020ELECTRA} implemented absolute position dependency. In contrast, models like XLNet \cite{NEURIPS2019_dc6a7e65}, DeBERTa \cite{he2021deberta}, and Music Transformer \cite{huang2018music} utilized relative position dependency. RoPE integrates both techniques by encoding the former with a rotation matrix and explicitly incorporating the latter in the self-attention formulation.

\section{Methodology}
\noindent In this section, we initially delineate the problem and present our hypotheses regarding its nature. Subsequently, we expound upon the methodology employed for the proposed frameworks and align them with the study objectives. The framework is delineated in two iterations: the conceptual and implementation frameworks. In addition, we investigate the constituent elements of these frameworks and preprocess the datasets preceding inference.

\subsection{\textbf{Problem Formulation}}
\noindent Our hypothesis proposes that when an expert addresses a query in a particular domain, they expend a level of cognitive effort that may diverge from that required for a Common question. This indicates a correlation between the model's domain expertise and inference overhead, manifested in the form of the time and computational resources consumed during the process. Formally, this relationship can be expressed as a 4-tuple:

\begin{equation}
\textit{O}=f(t, g, x, y, q_{\text{y}})
\end{equation}

\noindent where \textit{x} represents the prompt length, \textit{y} denotes the response length, \textit{O} signifies the inference overhead, \textit{t} denotes the inference time in seconds, \textit{g} stands for the maximum GPU usage, and $q_{\text{y}}$ indicates the quality of the response.

\subsection{\textbf{Proposed Framework}}
\noindent A framework to assess a large foundational model's comprehension of different domains is presented in Fig. 1. This is a three-dimensional representation of the problem domain, model size, and response control. In terms of the problem domain, we examine cybersecurity, medical, finance, and common questions. The model size includes Gemma-2B and Gemma-7B for tasks related to text generation inference. The response output is controlled, limited to 50 words, and unrestricted.

\begin{figure}[ht]
 \centering
 \includegraphics[scale=0.4]{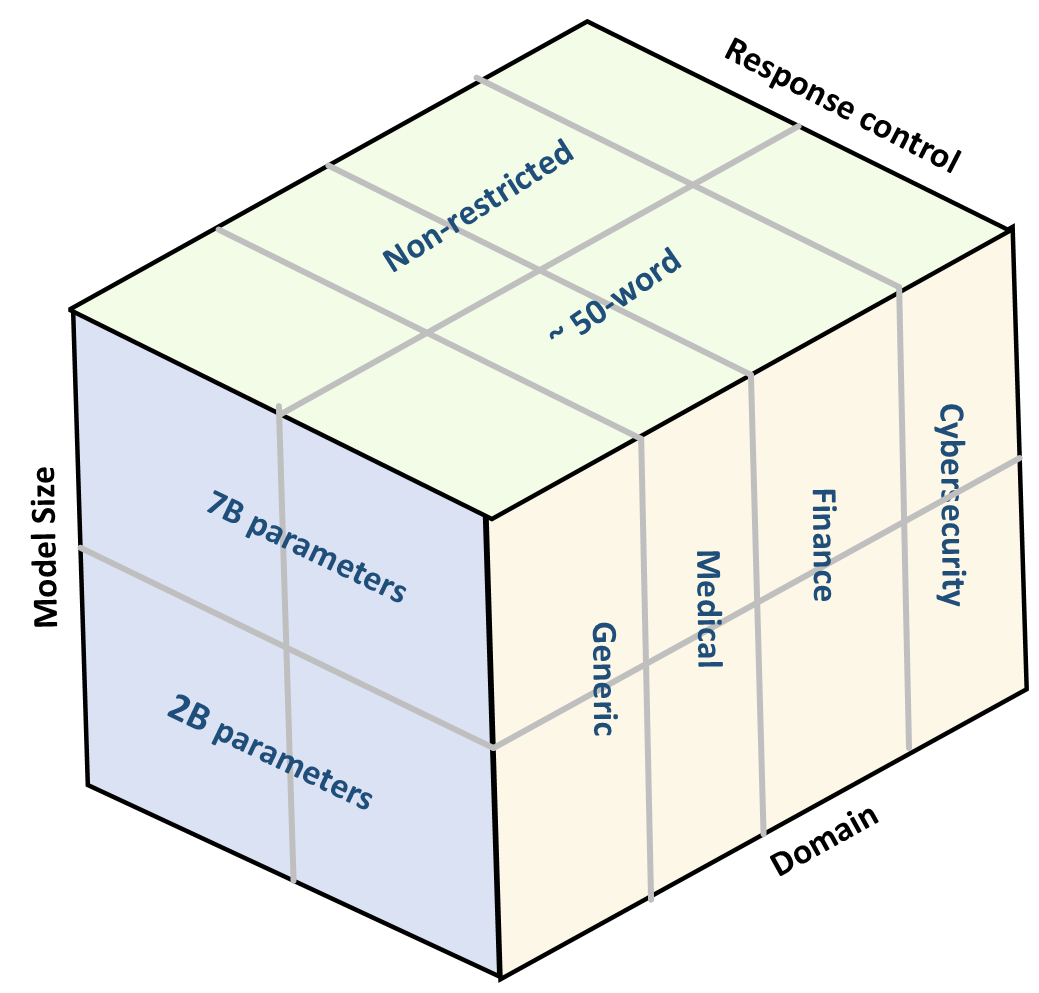}
 \caption{A framework for a large foundational model assessment about a domain understanding}
 \label{fig:image1010}
\end{figure}

\begin{figure*}[ht]
\centerline{\includegraphics[width=0.7\linewidth]{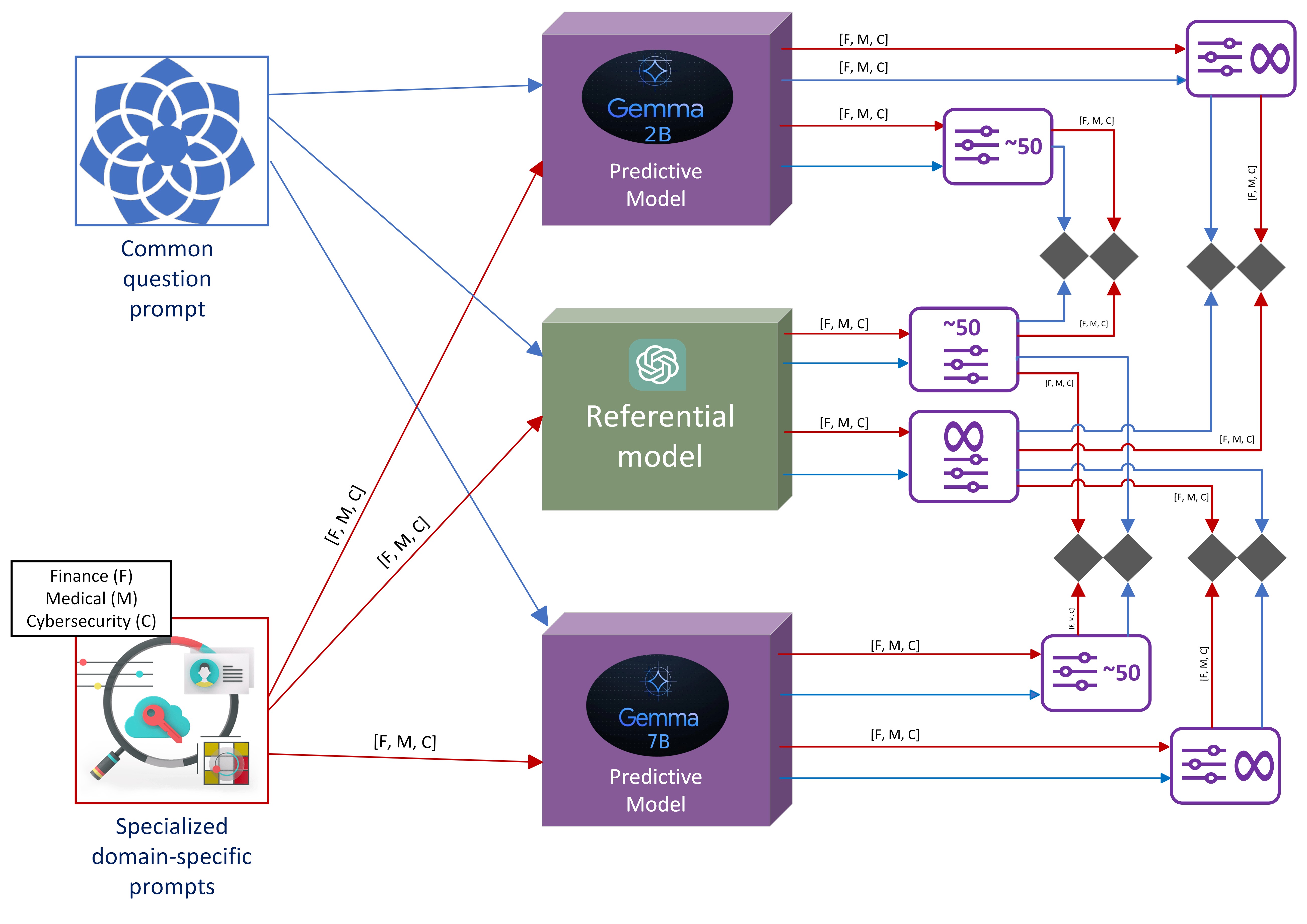}}
  \caption{A framework for the implementation of a large foundational model assessment about a domain understanding}
  \label{fig:multi_images}
\end{figure*}

We present the implementation framework in Fig. 2. Alongside the predictive models Gemma-2B and Gemma-7B, ChatGPT serves as a referential model against which we evaluate the quality of the predictive models.

To ensure precision, the experiment unfolded in sixteen distinct phases (4 x 4), each corresponding to an output line (blue and red which comprises F, M, and C) from the models as delineated in the framework. Pairing the output from each model, we configured each to produce responses containing 50 or unlimited words. Subsequently, we compared the predictive model output from each configuration with that of the referential model.

Throughout each inference phase, we meticulously recorded the inference time, response word length, GPU maximum consumption, and prompt word length data. In addition, we computed the inference throughput and latency and assessed response quality using the ROUGE-L and semantic text similarity (STS) metrics.

\subsection{\textbf{Data Analysis}}
In this study, we determine the statistical significance of the inference parameters and investigate the implications of the observed correlations between two variables. The correlation coefficient (typically Pearson's r) quantifies the linear relationship between two variables \(X\) and \(Y\) as follows:

\begin{equation}
  r = \frac{\sum (X_i - \bar{X})(Y_i - \bar{Y})}{\sqrt{\sum (X_i - \bar{X})^2 \sum (Y_i - \bar{Y})^2}}
\end{equation}

A value of \(r\) equal to 1 indicates perfect positive correlation, whereas a value of 1 indicates perfect negative correlation. A value of 0 implies no correlation.

\subsection{\textbf{Formulation of Outlier Technique}}
Given a set of data points that represent the correlation between two variables, we define upper and lower boundaries within which the slopes, \(m_{\text{min}}\) and \(m_{\text{min}}\), can be determined. The margin between these bounds represents the concentration area of the data points with upper and lower slope margins taken at a specific value derived from \(Eq. 5-11\). Outliers were identified as data points falling below the lower boundary.
 
\begin{equation}
m_{\text{central}} = \frac{y_{\text{2}}-y_{\text{1}}}{x_{\text{2}}-x_{\text{1}}}
\end{equation}
\begin{equation}
\theta_{\text{central}} = \arctan(m_{\text{central}})
\end{equation}
\begin{equation}
\theta_{\text{step}} = rad((\mu_{\text{x}} + (1.96* \sigma_{\text{x}}))\lambda)
\end{equation}
\begin{equation}
\theta_{\text{max}} = \theta_{\text{central}}+\theta_{\text{next}}
\end{equation}
\begin{equation}
\theta_{\text{min}} = \theta_{\text{central}}-\theta_{\text{next}}
\end{equation}
\begin{equation}
m_{\text{max}} = \tan(\theta_{\text{max}})
\end{equation}
\begin{equation}
m_{\text{min}} = \tan(\theta_{\text{min}})
\end{equation}

First, a straight line is plotted from (0, 0). The slope (\( m_{\text{central}} \)) and the angle in radians (\( \theta_{\text{central}} \)) are, respectively, calculated as follows. 5 and 6, respectively. The subsequent angle, \( \theta_{\text{step}} \), in radians, is computed in Eq. 7 using the 95\% confidence interval between the max-line and the central line and then to the min-line, where \(\lambda\) is the tuning parameter for angle adjustment, and \( \mu_{\text{x}} \) and \( \sigma_{\text{x}} \) are the mean and standard deviation of the interval, respectively. Furthermore, \( \theta_{\text{max}} \) (Eq. 8) and \( \theta_{\text{min}} \) (Eq. 9) are determined using \( \theta_{\text{central}} \) and \( \theta_{\text{step}} \), which are then used to compute \(m_{\text{min}}\) (Eq. 10) and \(m_{\text{min}}\) (Eq. 11), respectively.

\begin{figure*}
 \centering
 \begin{subfigure}[b]{0.23\textwidth}
 \centering{
 \includegraphics[width=\textwidth]{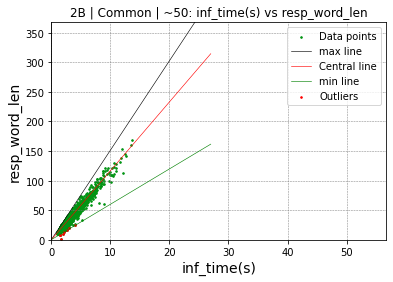} 
 \caption{\scriptsize \(2B/Common/\approx50\); \\ \(R=0.9761\); \(No. of Outlier=8\)} 
 \label{fig:image11}}
 \end{subfigure}
 \hspace{0.5em} 
 \begin{subfigure}[b]{0.23\textwidth}
 \centering
 \includegraphics[width=\textwidth]{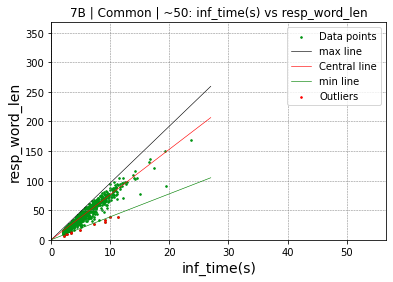} 
 \caption{\scriptsize \(7B/Common/\approx50\); \\ \(R=0.9585\);  \(No. of Outlier=11\)} 
 \label{fig:image2}
 \end{subfigure}
 \hspace{0.5em} 
 \begin{subfigure}[b]{0.23\textwidth}
 \centering
 \includegraphics[width=\textwidth]{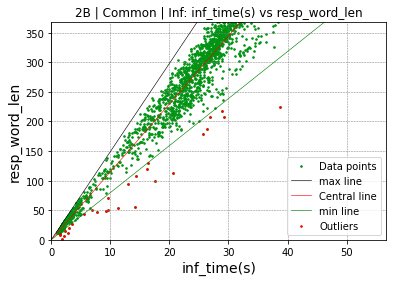} 
 \caption{\scriptsize \(2B/Common/\infty\); \\ \(R=0.9813\);  \(No. of Outlier=28\)} 
 \label{fig:image33}
 \end{subfigure}
 \hspace{0.5em} 
 \begin{subfigure}[b]{0.23\textwidth}
 \centering
 \includegraphics[width=\textwidth]{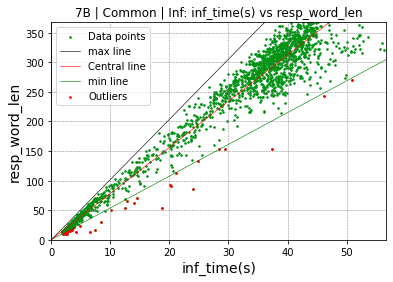} 
 \caption{\scriptsize \(7B/Common/\infty\); \\ \(R=0.9786\);  \(No. of Outlier=38\)} 
 \label{fig:image44}
 \end{subfigure}

 \vspace{1.2em} 

 \begin{subfigure}[b]{0.23\textwidth}
 \centering
 \includegraphics[width=\textwidth]{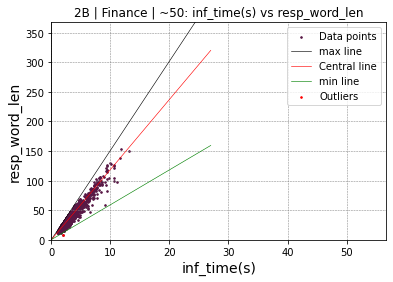} 
 \caption{\scriptsize \(2B/Finance/\approx50\); \\ \(R=0.9587\);  \(No. of Outlier=1\)} 
 \label{fig:image40}
 \end{subfigure}
 \hspace{0.5em} 
 \begin{subfigure}[b]{0.23\textwidth}
 \centering
 \includegraphics[width=\textwidth]{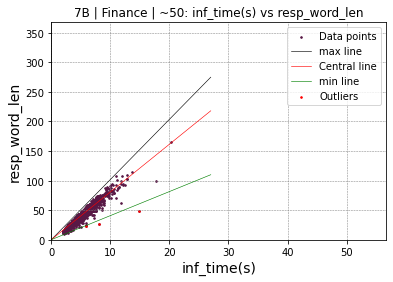} 
 \caption{\scriptsize \(7B/Finance/\approx50\); \\ \(R=0.9587\);  \(No. of Outlier=3\)} 
 \label{fig:image04}
 \end{subfigure}
 \hspace{0.5em} 
 \begin{subfigure}[b]{0.23\textwidth}
 \centering
 \includegraphics[width=\textwidth]{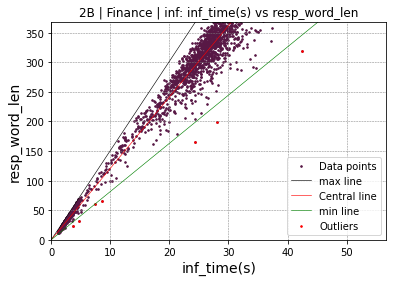} 
 \caption{\scriptsize  \(2B/Finance/\infty\); \\ \(R=0.9841\);  \(No. of Outlier=7\)} 
 \label{fig:image404}
 \end{subfigure}
 \hspace{0.5em} 
 \begin{subfigure}[b]{0.23\textwidth}
 \centering
 \includegraphics[width=\textwidth]{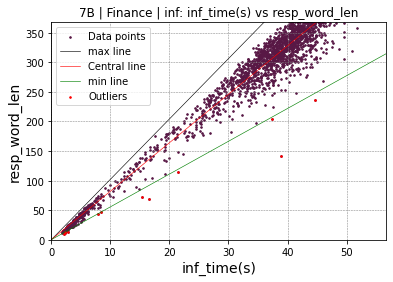} 
 \caption{\scriptsize  \(7B/Finance/\infty\); \\ \(R=0.9817\);  \(No. of Outlier=12\)} 
 \label{fig:image4_4}
\end{subfigure}

 \vspace{1.2em} 

 \begin{subfigure}[b]{0.23\textwidth}
 \centering
 \includegraphics[width=\textwidth]{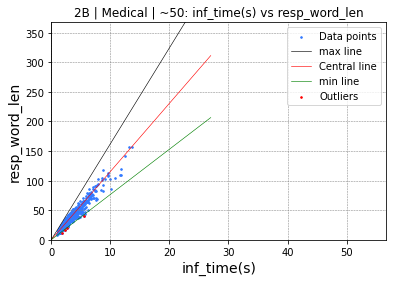} 
 \caption{\scriptsize  \(2B/Medical-health/\approx50\) \\ \(R=0.9654\);  \(No. of Outlier=4\)}  
 \label{fig:image303}
 \end{subfigure}
 \hspace{0.5em} 
 \begin{subfigure}[b]{0.23\textwidth}
 \centering
 \includegraphics[width=\textwidth]{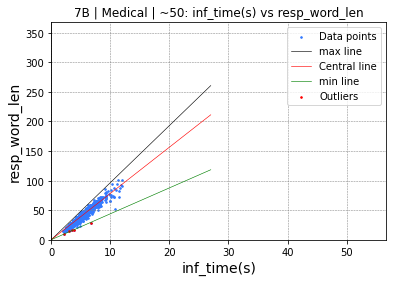} 
 \caption{\scriptsize \(7B/Medical-health/\approx50\); \\  \(R=0.9718\);  \(No. of Outlier=5\)}  
 \label{fig:image30}
 \end{subfigure}
 \hspace{0.5em} 
 \begin{subfigure}[b]{0.23\textwidth}
 \centering
 \includegraphics[width=\textwidth]{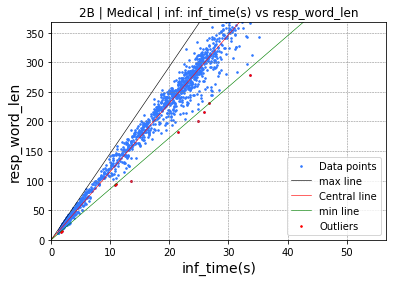} 
 \caption{\scriptsize \(2B/Medical-health/\infty\); \\ \(R=0.9931\);  \(No. of Outlier=10\)}  
 \label{fig:image333}
 \end{subfigure}
 \hspace{0.5em} 
 \begin{subfigure}[b]{0.23\textwidth}
 \centering
 \includegraphics[width=\textwidth]{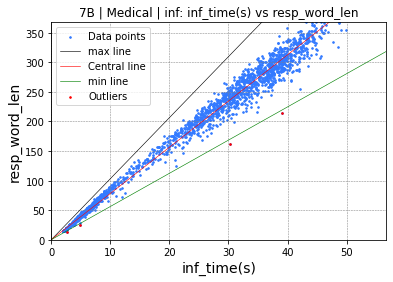} 
 \caption{\scriptsize \(7B/Medical-health/\infty\); \\ \(R=0.9937\); \(No. of Outlier=4\)}  
 \label{fig:image3}
 \end{subfigure}

 \vspace{1.2em} 

 \begin{subfigure}[b]{0.23\textwidth}
 \centering
 \includegraphics[width=\textwidth]{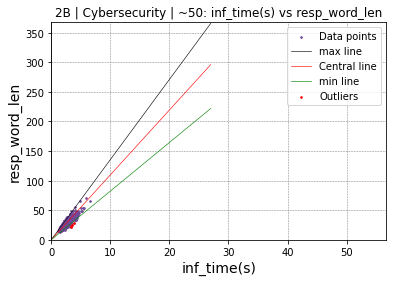} 
 \caption{\scriptsize \(2B/Cybersecury/\approx50\); \\  \(R=0.8568\);  \(No. of Outlier=4\)} 
 \label{fig:image5}
 \end{subfigure} 
 \hspace{0.5em} 
 \begin{subfigure}[b]{0.23\textwidth}
 \centering
 \includegraphics[width=\textwidth]{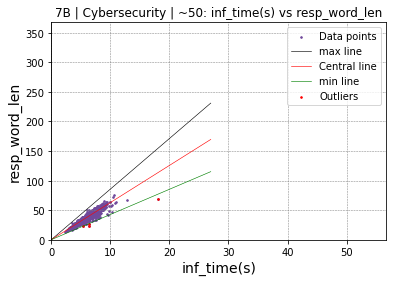} 
 \caption{\scriptsize \(7B/Cybersecury/\approx50\); \\ \(R=0.9368\);  \(No. of Outlier=4\)} 
 \label{fig:image6}
 \end{subfigure} 
 \hspace{0.5em} 
 \begin{subfigure}[b]{0.23\textwidth}
 \centering
 \includegraphics[width=\textwidth]{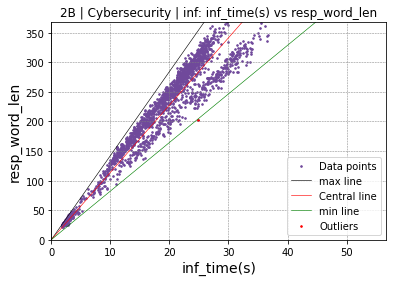} 
 \caption{\scriptsize \(2B/Common/\infty\); \\ \(R=0.8568\); \( No. of Outlier=2\)}
 \end{subfigure} 
 \hspace{0.5em} 
 \begin{subfigure}[b]{0.23\textwidth}
 \centering
 \includegraphics[width=\textwidth]{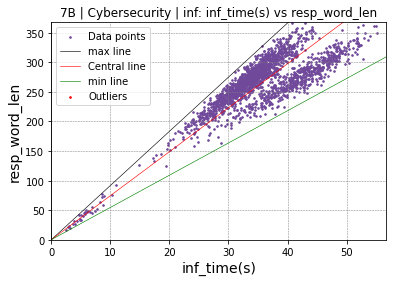} 
 \caption{\scriptsize \(7B/Common/\infty\); \\ \(R=0.9368\);  \(No. of Outlier=0\)} 
 \label{fig:image8}
 \end{subfigure} 
 \vspace{1.2em} 
\caption{Inference time (s) and response word length plots, estimating the correlation coefficient \((R)\), central line, upper and lower bounds, and outliers. The Common model had the highest number of outliers in all cases compared to the domain-specific responses.}
\label{fig:side_by_side}
\end{figure*}

\begin{table*}[ht]
\centering
\renewcommand{\arraystretch}{1.5} 
\caption{Outcomes of inference and ablation considering model size, domain, and response restriction.}
\resizebox{\textwidth}{!}{%
\label{tab:my-table}
\begin{tabular}{|
>{\columncolor[HTML]{FCE5CD}}c |
>{\columncolor[HTML]{FCE5CD}}l |
>{\columncolor[HTML]{FCE5CD}}c |c|c|c|c|cc|}
\hline
\cellcolor[HTML]{FCE5CD} & \multicolumn{1}{c|}{\cellcolor[HTML]{FCE5CD}}  & \cellcolor[HTML]{FCE5CD}  & \cellcolor[HTML]{C9DAF8}  & \cellcolor[HTML]{C9DAF8}  & \cellcolor[HTML]{C9DAF8} & \cellcolor[HTML]{C9DAF8} & \multicolumn{2}{c|}{\cellcolor[HTML]{C9DAF8}\textbf{Quality}}  \\ \cline{8-9} 
\multirow{-2}{*}{\cellcolor[HTML]{FCE5CD}\textbf{Model Size}} & \multicolumn{1}{c|}{\multirow{-2}{*}{\cellcolor[HTML]{FCE5CD}\textbf{Domain}}} & \multirow{-2}{*}{\cellcolor[HTML]{FCE5CD}\textbf{\begin{tabular}[c]{@{}c@{}}Response \\ Restriction\end{tabular}}} & \multirow{-2}{*}{\cellcolor[HTML]{C9DAF8}\textbf{\begin{tabular}[c]{@{}c@{}}Throughput\\ (\(\mu \pm \sigma\))\end{tabular}}} & \multirow{-2}{*}{\cellcolor[HTML]{C9DAF8}\textbf{\begin{tabular}[c]{@{}c@{}}Latency\\ (\(\mu \pm \sigma\))\end{tabular}}} & \multirow{-2}{*}{\cellcolor[HTML]{C9DAF8}\textbf{\begin{tabular}[c]{@{}c@{}}GPU Mem (MB)\\ (\(\mu \pm \sigma\)) \end{tabular}}} & \multirow{-2}{*}{\cellcolor[HTML]{C9DAF8}\textbf{\begin{tabular}[c]{@{}c@{}}Response Length\\ (\(\mu \pm \sigma\))\end{tabular}}} & \multicolumn{1}{c|}{\cellcolor[HTML]{C9DAF8}\textbf{\begin{tabular}[c]{@{}c@{}}ROUGE-L \\ (\(\mu \pm \sigma\))\end{tabular}}} & \cellcolor[HTML]{C9DAF8}\textbf{\begin{tabular}[c]{@{}c@{}}STS\\ (\(\mu \pm \sigma\))\end{tabular}} \\ \hline
2B & Common &  $\approx$50 & 12.0 ± 2.88  & 0.09 ± 0.08  & 2766 (4.43\%) & 35.31 ± 40.68  & \multicolumn{1}{c|}{0.25 ± 0.22} & 0.74± 0.38 \\ \hline
2B & Cybersecurity &  $\approx$50 & 11.26 ± 2.76 & 0.09 ± 0.02  & 3172 (5.08\%) & 25.72  ± 11.47 & \multicolumn{1}{c|}{0.29 ± 0.17} & 0.74± 0.25 \\ \hline
2B & Medical &  $\approx$50 & 12.08 ± 2.63 & 0.08 ± 0.02  & 2766 (4.43\%) & 28.94 ± 27.36  & \multicolumn{1}{c|}{0.26 ± 0.2}  & 0.72 ± 0.4 \\ \hline
2B & Finance &  $\approx$50 & 12.19 ± 2.77 & 0.08 ± 0.02  & 2770 (4.43\%) & 34.69 ± 36.16  & \multicolumn{1}{c|}{0.29 ± 0.2}  & 0.78 ± 0.27 \\ \hline
7B & Common &  $\approx$50 & 7.58 ± 1.75  & 0.13 ± 0.04  & 8760 (14.0\%) & 35.07 ± 33.51  & \multicolumn{1}{c|}{0.30 ± 0.25} & 0.78± 0.36 \\ \hline
7B & Cybersecurity &  $\approx$50 & 6.29 ± 0.95  & 0.16 ± 0.03  & 8906 (14.26\%) & 35.71 ± 17.13  & \multicolumn{1}{c|}{0.34 ± 0.23} & 0.68 ± 0.26 \\ \hline
7B & Medical &  $\approx$50 & 7.73 ± 1.22  & 0.13 ± 0.02  & 7121 (14.01\%) & 37.71 ± 26.29  & \multicolumn{1}{c|}{0.32 ± 0.2}  & 0.82 ± 0.25 \\ \hline
7B & Finance &  $\approx$50 & 7.98 ± 1.65  & 0.13 ± 0.03  & 8750 (14.0\%) & 38.41 ± 31.55  & \multicolumn{1}{c|}{0.33 ± 0.21} & 0.83 ± 0.21 \\ \hline
2B & Common & \(\infty\) & 11.79 ± 2.8  & 0.09 ± 0.08  & 2792 (4.47\%) & 207.75 ± 251.57 & \multicolumn{1}{c|}{0.19 ± 0.19} & 0.69 ± 0.49 \\ \hline
2B & Cybersecurity & \(\infty\) & 11.91 ± 2.44 & 0.13 ± 0.03  & 3564 (5.71\%) & 268.27 ± 91.39 & \multicolumn{1}{c|}{0.21 ± 0.12} & 0.72 ± 0.26 \\ \hline
2B & Medical & \(\infty\) & 12.37 ± 2.28 & 0.08 ± 0.02  & 2792 (4.47\%) & 128.23 ± 220.98 & \multicolumn{1}{c|}{0.21 ± 0.19} & 0.71 ± 0.39 \\ \hline
2B & Finance & \(\infty\) & 12.33 ± 2.15 & 0.08 ± 0.02  & 2790 (4.47\%) & 248.56 ± 248.27 & \multicolumn{1}{c|}{0.21 ± 0.13} & 0.78 ± 0.22 \\ \hline
7B & Common & \(\infty\) & 7.86 ± 1.83  & 0.13 ± 0.04  & 8800 (14.09\%) & 215.5 ± 247.28 & \multicolumn{1}{c|}{0.21 ± 0.22} & 0.72 ±  0.5 \\ \hline
7B & Cybersecurity & \(\infty\) & 7.61 ± 1.56  & 0.09 ± 0.02  & 8764 (14.03\%) & 206.79 ± 186.24 & \multicolumn{1}{c|}{0.21 ± 0.12} & 0.72 ± 0.26 \\ \hline
7B & Medical & \(\infty\) & 7.9 ± 1.11 & 0.13 ± 0.02  & 71214 (14.0\%) & 176.46 ± 222.19 & \multicolumn{1}{c|}{0.21 ± 0.19} & 0.8 ± 0.24 \\ \hline
7B & Finance & \(\infty\) & 8.16 ± 1.4 & 0.12 ± 0.03  & 8750 (14.0\%) & 265.9 ± 223.71 & \multicolumn{1}{c|}{0.23 ± 0.14} & 0.81 ± 0.2 \\ \hline
\end{tabular}%
}
\end{table*}

\subsection{\textbf{Dataset}} 
\noindent In the investigation, we examined two categories of datasets: Common and domain-specific datasets, each comprising 2019 instances. The Common dataset was retrieved from GLUE (General Language Understanding Evaluation) \cite{wang2018glue}. To ensure fairness, we endeavored to exclude instances containing toxic content, as they could potentially be rejected by the model, affecting the response length. Furthermore, we ensured that the prompts included commonplace questions that anyone could easily answer without particular expertise.

Furthermore, the domain-specific datasets consist of three domains: cybersecurity-oriented, finance-oriented, and medical-oriented datasets. The cybersecurity dataset was obtained from \cite{fayyazi2023advancing}. Primarily, it consists of attack procedures originally curated from the MITRE ATT\&CK \cite{mitre_attack}, a globally accessible knowledge base of adversary tactics and techniques derived from real-world observations. Given an attack procedure as a prompt, the underlying premise is that we anticipate the model's ability to predict what the attacker aims to achieve. The finance-oriented dataset was acquired from \cite{gaurang_bharti_2024}, originally combining Stanford's Alpaca and FiQA datasets which have been used to facilitate the training and fine-tuning of diverse models tailored for financial applications. Subsequently, the medical-oriented dataset sourced from \cite{aimedicaldataset} was employed by the original author for training purposes in the development of the AI medical chatbot.

\noindent In the investigation, we examined two categories of datasets: Common and domain-specific datasets, each comprising 2019 instances. The Common dataset was retrieved from GLUE (General Language Understanding Evaluation) \cite{wang2018glue}. To ensure fairness, we excluded instances containing toxic content because they could be rejected by the model, which affected the response length.

Furthermore, we ensured that the prompts included commonplace questions that anyone could easily answer without particular expertise. In addition, the domain-specific datasets comprise three domains: cybersecurity-oriented, finance-oriented, and medical-oriented datasets. The cybersecurity dataset was obtained from \cite{fayyazi2023advancing}. Primarily, it consists of attack procedures originally curated from the MITER ATT\&CK \cite{mitre_attack}, a globally accessible knowledge base of adversary tactics and techniques derived from real-world observations. Given an attack procedure as a prompt, the underlying premise is that we anticipate the model's ability to predict what the attacker intends to achieve. The finance-oriented dataset was acquired from \cite{gaurang_bharti_2024}, originally combining Stanford's Alpaca and FiQA datasets, which have been used to facilitate the training and fine-tuning of diverse models tailored for financial applications. Subsequently, a medical-oriented dataset was sourced from \cite{aimedicaldataset}. The original author used the same dataset to train AI-medical chatbots.

\begin{table*}[ht]
\centering
\renewcommand{\arraystretch}{1.5} 
\caption{Outlier analysis (Figure 5. The values highlighted in red indicate that the outliers are below the overall values specified in Table I, and the values in blue indicate that they exceed the standard values.}
\label{tab:my-table1}
\resizebox{\textwidth}{!}{%
\begin{tabular}{|
>{\columncolor[HTML]{FCE5CD}}c |
>{\columncolor[HTML]{FCE5CD}}c |
>{\columncolor[HTML]{FCE5CD}}c |c|c|c|c|c|c|c|c|c|c|c|c|c|c|c|}
\hline
\textbf{\begin{tabular}[c]{@{}c@{}}Model \\ Size\end{tabular}} & \textbf{Domain} & \textbf{\begin{tabular}[c]{@{}c@{}}Response \\ Restriction\end{tabular}} & \cellcolor[HTML]{C9DAF8}\textbf{\begin{tabular}[c]{@{}c@{}}Inf\_time-\\ response\_len \\ Correlation\\ (\(R\))\end{tabular}} & \cellcolor[HTML]{C9DAF8}\textbf{\begin{tabular}[c]{@{}c@{}}Max \\ Slope \\ (\(m_{\text{max}}\))\end{tabular}} & \cellcolor[HTML]{C9DAF8}\textbf{\begin{tabular}[c]{@{}c@{}}Central \\ Slope\\ (\(m_{\text{central}}\)) \end{tabular}} & \cellcolor[HTML]{C9DAF8}\textbf{\begin{tabular}[c]{@{}c@{}}Min\\ Slope \\ (\(m_{\text{min}}\))\end{tabular}} & \cellcolor[HTML]{C9DAF8}\textbf{\begin{tabular}[c]{@{}c@{}}Max \\ angle \\in rad \\ (\( \theta_{\text{max}} \)) \end{tabular}} & \cellcolor[HTML]{C9DAF8}\textbf{\begin{tabular}[c]{@{}c@{}}Central \\ angle in \\ rad \\ (\( \theta_{\text{central}} \)) \end{tabular}} & \cellcolor[HTML]{C9DAF8}\textbf{\begin{tabular}[c]{@{}c@{}}Min \\ angle in \\ rad \\ (\(\theta_{\text{min}}\)) \end{tabular}} & \cellcolor[HTML]{C9DAF8}\textbf{\begin{tabular}[c]{@{}c@{}}No of \\ Outlier\end{tabular}} & \cellcolor[HTML]{C9DAF8}\textbf{\begin{tabular}[c]{@{}c@{}}Inf\_time(s)\\ (\(\mu \pm \sigma\))\end{tabular}} & \cellcolor[HTML]{C9DAF8}\textbf{\begin{tabular}[c]{@{}c@{}}resp\_word\_len\\ (\(\mu \pm \sigma\))\end{tabular}} & \cellcolor[HTML]{C9DAF8}\textbf{\begin{tabular}[c]{@{}c@{}}prompt\_word \\ \_len\\ (\(\mu \pm \sigma\))\end{tabular}} & \cellcolor[HTML]{C9DAF8}\textbf{\begin{tabular}[c]{@{}c@{}}Latency\\ (\(\mu \pm \sigma\))\end{tabular}} & \cellcolor[HTML]{C9DAF8}\textbf{\begin{tabular}[c]{@{}c@{}}Throughput\\ (\(\mu \pm \sigma\))\end{tabular}} & \cellcolor[HTML]{C9DAF8}\textbf{\begin{tabular}[c]{@{}c@{}}ROUGE-L\\ (\(\mu \pm \sigma\))\end{tabular}} & \cellcolor[HTML]{C9DAF8}\textbf{\begin{tabular}[c]{@{}c@{}}STS\\ (\(\mu \pm \sigma\))\end{tabular}} \\ \hline
2B & Common & $\approx$50 & 0.9761 & 15.14 & 11.66 & 5.99 & 1.50 & 1.49  & 1.41 & 8 & {\color[HTML]{980000} 2.3±0.9}  & {\color[HTML]{980000} 12.5±7.3} & {\color[HTML]{980000} 17.1±2.2} & {\color[HTML]{0000FF} 0.4±0.5} & {\color[HTML]{980000} 5.2±1.9} & {\color[HTML]{0000FF} 0.3±0.2} & {\color[HTML]{980000} 0.7±0.2} \\ \hline
2B & Cybersecurity &  $\approx$50 & 0.8568 & 13.56 & 10.97 & 8.22 & 1.51 & 1.48  & 1.45 & 4 & {\color[HTML]{0000FF} 3.46±0.29} & {\color[HTML]{980000} 24.50±2.65}  & {\color[HTML]{980000} 38.75±4.86} & {\color[HTML]{0000FF} 0.14±0.00} & {\color[HTML]{980000} 7.08±0.21} & {\color[HTML]{980000} 0.26±0.07} & {\color[HTML]{980000} 0.73±0.12} \\ \hline
2B & Medical &  $\approx$50 & 0.9654 & 16.23 & 11.54 & 7.65 & 1.51 & 1.48  & 1.44 & 4 & {\color[HTML]{0000FF} 3.06±1.69} & {\color[HTML]{980000} 22.00±12.36} & {\color[HTML]{980000} 18.75±1.71} & {\color[HTML]{0000FF} 0.14±0.00} & {\color[HTML]{980000} 7.17±0.14} & {\color[HTML]{980000} 0.25±0.12} & {\color[HTML]{0000FF} 0.84±0.04} \\ \hline
2B & Finance &  $\approx$50 & 0.9674 & 15.1  & 11.87 & 5.91 & 1.50 & 1.49  & 1.4  & 1 & {\color[HTML]{980000} 1.99±0.0} & {\color[HTML]{980000} 8±0.0}  & {\color[HTML]{0000FF} 31±0.0} & {\color[HTML]{0000FF} 0.25±0.0}  & {\color[HTML]{980000} 4.03±0.0}  & {\color[HTML]{980000} 0.06±0.0}  & {\color[HTML]{980000} 0.52±0.0}  \\ \hline
7B & Common &  $\approx$50 & 0.9585 & 9.61  & 7.65  & 3.88 & 1.47 & 1.44  & 1.32 & 11  & {\color[HTML]{0000FF} 5.97±3.41} & {\color[HTML]{980000} 21.09±12.32} & {\color[HTML]{980000} 17.55±1.86} & {\color[HTML]{0000FF} 0.29±0.02} & {\color[HTML]{980000} 3.51±0.23} & {\color[HTML]{980000} 0.06±0.05} & {\color[HTML]{980000} 0.08±0.11} \\ \hline
7B & Cybersecurity &  $\approx$50 & 0.9368 & 8.56  & 6.28  & 4.26 & 1.45 & 1.41  & 1.34 & 4 & {\color[HTML]{0000FF} 9.02±6.00} & {\color[HTML]{980000} 35.50±22.41} & {\color[HTML]{0000FF} 40.75±6.18} & {\color[HTML]{0000FF} 0.25±0.02} & {\color[HTML]{980000} 3.99±0.31} & {\color[HTML]{980000} 0.19±0.12} & {\color[HTML]{980000} 0.59±0.17} \\ \hline
7B & Medical &  $\approx$50 & 0.9718 & 9.66  & 7.83  & 4.38 & 1.47 & 1.44  & 1.35 & 5 & {\color[HTML]{980000} 3.83±1.73} & {\color[HTML]{980000} 16.80±6.72}  & {\color[HTML]{0000FF} 20.40±1.82} & {\color[HTML]{0000FF} 0.22±0.01} & {\color[HTML]{980000} 4.46±0.25} & {\color[HTML]{980000} 0.24±0.05} & {\color[HTML]{980000} 0.83±0.05} \\ \hline
7B & Finance &  $\approx$50 & 0.9587 & 10.18 & 8.08  & 4.07 & 1.47 & 1.45  & 1.33 & 3 & {\color[HTML]{0000FF} 9.61±3.79} & {\color[HTML]{980000} 33.00±11.43} & {\color[HTML]{0000FF} 20.67±3.86} & {\color[HTML]{0000FF} 0.29±0.02} & {\color[HTML]{980000} 3.51±0.27} & {\color[HTML]{0000FF} 0.36±0.07} & {\color[HTML]{0000FF} 0.86±0.03} \\ \hline
2B & Common & \(\infty\) & 0.9813 & 14.91 & 11.49 & 7.97 & 1.5 & 1.48  & 1.45 & 28  & {\color[HTML]{980000} 13.09±10.34} & {\color[HTML]{980000} 85.89±72.66} & {\color[HTML]{0000FF} 11.71±4.85} & {\color[HTML]{0000FF} 0.23±0.32} & {\color[HTML]{980000} 6.15±1.69} & {\color[HTML]{0000FF} 0.27±0.16} & {\color[HTML]{0000FF} 0.73±0.19} \\ \hline
2B & Cybersecurity & \(\infty\) & 0.8568 & 14.22 & 11.39 & 8.22 & 1.5 & 1.48  & 1.45 & 2 & {\color[HTML]{0000FF} 65.79±57.86} & {\color[HTML]{980000} 225.00±32.53} & {\color[HTML]{980000} 32.00±4.24} & {\color[HTML]{0000FF} 0.28±0.22} & {\color[HTML]{980000} 5.22±4.10} & {\color[HTML]{980000} 0.04±0.05} & {\color[HTML]{980000} 0.04±0.01} \\ \hline
2B & Medical & \(\infty\) & 0.9931 & 14.65 & 11.55 & 8.62 & 1.5 & 1.48  & 1.46 & 10  & {\color[HTML]{0000FF} 17.13±10.97} & {\color[HTML]{0000FF} 142.30±92.38} & 13.50±4.35  & {\color[HTML]{0000FF} 0.12±0.01} & {\color[HTML]{980000} 8.22±0.41} & {\color[HTML]{0000FF} 0.24±0.06} & {\color[HTML]{0000FF} 0.78±0.12} \\ \hline
2B & Finance & \(\infty\) & 0.9841 & 15.1  & 12.8  & 8.15 & 1.5 & 1.49  & 1.45 & 7 & {\color[HTML]{980000} 17.01±14.83} & {\color[HTML]{980000} 123.43±109.33}  & {\color[HTML]{0000FF} 15.29±7.02} & {\color[HTML]{0000FF} 0.14±0.01} & {\color[HTML]{980000} 7.19±0.62} & {\color[HTML]{980000} 0.15±0.04} & {\color[HTML]{980000} 0.61±0.14} \\ \hline
7B & Common & \(\infty\) & 0.9786 & 10.18 & 7.81  & 5.38 & 1.47 & 1.44  & 1.39 & 38  & {\color[HTML]{980000} 13.60±14.72} & {\color[HTML]{980000} 64.08±76.32} & {\color[HTML]{980000} 11.42±3.84} & {\color[HTML]{0000FF} 0.23±0.07} & {\color[HTML]{980000} 4.52±0.89} & {\color[HTML]{980000} 0.31±0.19} & {\color[HTML]{980000} 0.76±0.17} \\ \hline
7B & Cybersecurity & \(\infty\) & 0.9368 & 9.17  & 7.44  & 5.45 & 1.46 & 1.44  & 1.39 & 0 & -  & - & -  & {\color[HTML]{0000FF} -} & - & - & -  \\ \hline
7B & Medical & \(\infty\) & 0.9937 & 10.32 & 7.82  & 5.62 & 1.47 & 1.44  & 1.39 & 4 & {\color[HTML]{980000} 19.19±18.19} & {\color[HTML]{980000} 103.75±100.34}  & {\color[HTML]{980000} 13.00±3.16} & {\color[HTML]{0000FF} 0.19±0.01} & {\color[HTML]{980000} 5.22±0.26} & {\color[HTML]{0000FF} 0.25±0.05} & 0.72±0.12 \\ \hline
7B & Finance & \(\infty\) & 0.9817 & 10.21 & 8.2 & 5.55 & 1.47 & 1.45  & 1.39 & 12  & {\color[HTML]{980000} 16.68±15.63} & {\color[HTML]{980000} 81.00±77.58} & {\color[HTML]{0000FF} 14.50±5.20} & {\color[HTML]{0000FF} 0.21±0.03} & {\color[HTML]{980000} 4.87±0.63} & {\color[HTML]{980000} 0.05±0.04} & {\color[HTML]{980000} 0.06±0.05} \\ \hline
\end{tabular}%
}
\end{table*}

\section{Result Discussion}
\subsection{\textbf{Analysis of Response}}
\noindent 
The study results are presented in Table 1. Common prompts tend to generate responses with greater diversity in length and significantly longer inference times. The 7B models consume more GPU memory than their 2B counterparts, which is expected due to the differences in size. In all cases, the 2B models achieved higher throughputs, over 50\%, than the 7B models, indicating that the 7B models incur more computation and memory overhead than the 2B models.

Regarding quality, using semantic textual similarity (STS) with ChatGPT responses as a reference model, the top five highest scores for semantic textual similarity (STS) are: \(7B/Finance/\approx50\), \(7B/Medical/\approx50\), \(7B/Finance/\infty\), \(7B/Medical/\infty\), and \(7B/Common/\approx50\). Conversely, \(2B/Cybersecurity/\infty\), \(7B/Cybersecurity/\infty\), and \(7B/Cybersecurity/\approx50\) have the lowest STS. This alignment implies that the words and phrases in \(7B/Common/\approx50\) convey meanings that are more similar to those conveyed in the reference ChatGPT text. This indicates a higher degree of agreement or relevance. Higher STS scores indicate greater semantic similarity 
(or word-level similarity) between the text and reference text. In addition, restricted responses showed better ROUGE-L values for both common and domain-specific prompts, which implies that the responses contain similar keywords as the reference responses. It is important to note that the quality assessment of the responses using the STS and ROUGE-L in this study may not reflect the true value of response quality because the predicted and reference responses are often not of equal length and differ with a wide margin, which is a major criterion that may affect the quality assessment of datasets.

\subsection{\textbf{Analysis of Correlation and Outliers}}
\noindent 
Our results demonstrate a significant correlation between inference time and response length (Figure 5a-p). Table II provides further insight into this relationship. The 7B and 2B models with unrestricted responses to Medical, Finance, and Common prompts exhibit the highest time-response correlations, and the cybersecurity model exhibits the lowest time-response correlation across all scenarios. Several key observations emerge from the results. The variability in correlation is depicted in Fig. 5 can be computed using the previously discussed outliers technique (Section III -subsection C). A higher response in a shorter time indicates high throughput, and the primary objective is to identify outliers on the x-axis, i.e., inf-time. For the response-restricted cases, the value of \(\lambda\) is set to \(0.005\) for \( \theta_{\text{max}} \) and \(0.5\) for \( \theta_{\text{min}} \), whereas, for the unrestricted cases, the value of \(\lambda\) is set to \(0.0005\) for \( \theta_{\text{max}} \) and \(0.05\) for \( \theta_{\text{min}} \). This resulted in a variability ratio of \(10:1\) between restricted and unrestricted response lengths.


Common prompts exhibited the highest number of outliers across all cases compared to domain-specific prompts. This finding further confirms the inconsistency in response length associated with common prompts relative to their domain-specific counterparts. Within the domain-specific category, \(7B/Finance/\infty\) manifests the highest number of outliers, which is attributed to its inclusion of content from related fields such as insurance, accounting, and taxation. In contrast, cybersecurity and medical prompts consistently yielded concise responses in all cases. 
The analysis of the outliers indicates that the values are nearly uniformly below the overall mean across all parameters except latency.

\section{Conclusion}
This study investigates the inference behavior of foundation models of varying sizes under common and domain-specific prompts, such as those related to cybersecurity, medical, and finance domains. This study examines these behaviors under conditions in which response lengths are both restricted and unrestricted. We present a framework to assess large foundational models in terms of domain understanding and outlier formulation. The results indicate that model size and types of prompts used for inference significantly influence response length and quality, as larger datasets for training provide more information across various domains. In addition, common prompts, which include different types of queries, generate diverse responses and may result in inconsistent response lengths when the same prompt is used multiple times or in different ways. In contrast, domain-specific prompts consistently generate concise responses. Therefore, we recommend eliminating irrelevant domains in the language model information prior to fine-tuning domain-specific tasks. For example, when provided with sufficient datasets to fine-tune the 2 billion parameter model for a cybersecurity downstream task, we advocate for a full fine-tuning approach rather than employing a parameter-efficient technique. This method involves the elimination of irrelevant domains, allowing the target domain to become predominant. Such an approach preserves the model's size while facilitating the generation of concise and consistent responses in a minimal amount of time.

While this study aimed to focus on resource utilization, response length, and response quality, we did not observe significant differences in resource utilization and response quality assessment because the results did not show statistically significant differences among the cases. Consequently, future research should focus on a comprehensive investigation of response quality across various domains and determine whether response length correlates with quality. In addition, resource usage must be assessed in a manner that does not affect inference time.

\bibliographystyle{IEEEtran}
\bibliography{main} 

\end{document}